\documentclass[twocolumn]{article}
\usepackage[width=17cm,height=22cm]{geometry}
\usepackage[utf8]{inputenc}
\usepackage{fancyvrb}
\usepackage{authblk}
\usepackage{hyperref}
\usepackage{color}
\usepackage{amsmath}
\usepackage[T1]{fontenc}

\usepackage{xcolor}
\hypersetup{
    colorlinks,
    linkcolor={black},
    citecolor={black},
    urlcolor={black}
}

\title{Predicting CEFRL levels in learner English on the basis of metrics and full texts}
\date{}
\author[1]{Taylor Arnold}
\author[2]{Nicolas Ballier}
\author[3]{Thomas Gaillat}
\author[4]{Paula Liss\'on}
\affil[1]{University of Richmond, USA\thanks{taylor.arnold@acm.org}}
\affil[2]{Universit\'e Paris Diderot, France}
\affil[3]{The Insight Centre for Data Analytics NUI Galway, Ireland}
\affil[4]{Universit\'e Paris Diderot, France}

\begin{document}
\maketitle

\begin{abstract}
This paper analyses the contribution of language metrics and, potentially, of linguistic structures, to classify French learners of English according to levels of the Common European Framework of Reference for Languages (CEFRL). The purpose is to build a model for the prediction of learner levels as a function of language complexity features. We used the EFCAMDAT corpus \cite{geertzen2013automatic}, a database of one million written assignments by learners. After applying language complexity metrics on the texts, we built a representation matching the language metrics of the texts to their assigned CEFRL levels. Lexical and syntactic metrics were computed with LCA and LSA \cite{lu_computational_2014} and koRpus \cite{michalke2017}. Several supervised learning models were built by using Gradient Boosted Trees and Keras Neural Network methods and by contrasting pairs of CEFRL levels. Results show that it is possible to implement pairwise distinctions, especially for levels ranging from A1 to B1 (A1=>A2: 0.916 AUC and A2=>B1: 0.904 AUC). Model explanation reveals significant linguistic features for the predictiveness in the corpus. Word tokens and word types appear to play a significant role in determining levels. This shows that levels are highly dependent on specific semantic profiles. 



\end{abstract}

\medskip


\section{Introduction}
This paper focuses on the detection of language levels in Second Language Acquisition. Foreign language education centers provide courses that are tailored according to the different levels of their learners and this leads to two requirements. In the process of learning, it is paramount to provide regular evaluations to both learners and teachers so as to help them focus on specific areas to train upon. There is also a growing demand to group learners homogeneously in order to set adequate teaching objectives and methods. These two requirements rely on language assessment tests whose design and organization are labor intensive and thus costly. Currently, language centers rely on instructors to design and manually correct tests. Alternatively, they use specifically designed short-context and rule-based on-line exercises in which a discrete number of specific language errors is used as a paradigm for level assignment. This creates a bias to use errors as the sole criterion for assessment. Recent research has shown that metrics used in the domain of text mining and NLP can help characterize complexity and thus levels. Consequently, there is a need to use error independent tools to compute levels. 
\par
The objective of our experiment is to show that a supervised learning approach is possible. By creating a vector representation of texts matched with language complexity metrics, we build a predictive model for language levels. We use the EFCAMDAT \footnote{The training and test data were selected and manipulated independently of direct involvement from the EF and Cambridge research teams.This corpus is publicly available at https://corpus.mml.cam.ac.uk/efcamdat2/public\_html/} corpus \cite{geertzen2013automatic} in which learner texts are classified according to the six levels of the Common European Framework of Reference for Languages (CEFRL). The corpus is used to create a dataset made up of complexity metrics. These complexity metrics are used as input for a Neural Network (NN) whose output layer consists of CEFRL levels. The rest of this paper is organized as follows. Section \ref{RElated_work} covers previous work in the domain of automatic language proficiency assignment. In Section \ref{method}, we present the corpus and the method used to build the NN model. Section \ref{sect:models} describes the models and we discuss results and conclude in Section \ref{Sect:discuss-conclude}





\section{Related Work} \label{RElated_work}
There is a large body of research in automatic language scoring starting with \cite{page_use_1968}. Over the last four decades, there has been various methods to provide language level analysis. Methods evolved from rule-based approaches focused on pattern matching to Machine Learning (ML) approaches including unsupervised and supervised learning methods.  
\par
Rule-based systems have included analyses relying on error detection \cite{leacock_c-rater:_2003,mitchell_towards_2002} while other systems identify features assumed to be to be indicative of proficiency \cite{higgins_three-stage_2011}. With the advent of ML techniques, probabilistic models have appeared. Some approaches use unsupervised learning \cite{tono_automatic_2013} but, since the task of score assignment relies on human annotated corpora, most ML strategies rely on supervised learning with SVM \cite{yoon_assessment_2012} or linear regression and decision tree \cite{chen_computing_2011}. Our proposal falls into this category but it uses a Neural Network including several LTSM and dense layers. The input layer includes a multi-dimensional feature representation of written essays and the output corresponds to language levels. 
\par
This area of research is closely linked to the research on criterial features that define levels. All the aforementioned studies include a large number of features based on morpho-syntactic patterns, word counts, text and readability metrics \cite{Vajjala2018}. Authors tested their significance in terms of correlation or classification performance. Another perspective is to specifically focus on weighing feature significance, e.g. by applying strategies based on entropy \cite{flanagan_relationship_2015}, errors \cite{tono_automatic_2013} or lexical metrics \cite{ballier_classifying_2016}. In our experiment, we use lexical and syntactic complexity features \cite{lu_automatic_2010,lu_relationship_2012}. Our approach supports two perspectives, i.e. classification according to features and modeling for feature significance analysis with a gradient boost tree model. 
\par
Few studies make use of the CERFL as a standard for level description. In English, many studies use other language level scales such as \cite{flanagan_relationship_2015} with the NICT-JLE corpus levels, \cite{crossley_predicting_2011} with  TOEFL levels and \cite{yannakoudakis_new_2011} for the Cambridge Learner Corpus (CLC). We built our data set with texts from the EF-Cambridge Open Language Database (EFCamDat) as a Gold Standard and conducted experiments on English texts classified according to the CEFRL.


\section{Method} \label{method}

\subsection{Data extraction}
All the available raw texts from French learners were downloaded from the EFCAMDAT database in separate XML files. Each XML file corresponded to the EFCAMDAT levels associated with each one of the CEFRL levels. In total, 
41,626 texts (approx. 3,298,343 tokens), corresponding to 128 units and 7,695 French learners were downloaded.



\subsection{Lexical Diversity Metrics}
Most of the lexical diversity metrics are based on the relationship between the numbers of types and tokens within a given text. That is, for example, the case of the 
\textbf{TTR:} (types/tokens), and some of its mathematical transformations, such as the \textbf{MSTTR} (types/tokens, with fragments of \textit{n} tokens), \textbf{Herdan’s C} (logTypes/logTokens), \textbf{Guiraud’s RTTR} (types/$\sqrt{tokens}$), \textbf{Carrol’s CTTR} (types/ $2\sqrt{tokens}$), \textbf{Dugast’s Uber Index} ($log{Tokens}^2$ / logTypes - log tokens), \textbf{Summer’s Index (S)} (log(logTypes) / log(logTokens)), \textbf{Maas a:} $a^2 = (logTypes / logTokens) / logTokens^2$, \textbf{Maas log} ($logTypes_0$= $logTypes$/ $\sqrt{1- logTypes^2/logTokens}$), and \textbf{Yule’s K} ($10^4(\sum(fX*X^2) - tokens) / tokens^2$, where \textit{X}
is a vector with the frequencies of each type, and $f_X$ is the frequency for each \textit{X}). 

However, most of these metrics are said to be unreliable because they are highly dependent on text length \cite{tweedie1998variable}. As a consequence, a second generation of metrics, with more complex mathematical transformations, has been developed: \textbf{MTLD:} (types/factors, where factors are segments that have reached the stabilization point of TTR; \textbf{MATTR} (mean of moving TTR, generated through a 'window' technique of variable sizes that computes TTR of samples of the text); 
\textbf{MTLD-MA}, which combines both factors  and the window technique, or the \textbf{HDD-D} metric, which computes, for each type, the probability of finding any of its tokens in a random sample of 42 words taken from the text.

\subsection{Complexity Metrics}
Complexity metrics were generated using Xiaofei Lu's software \cite{lu_relationship_2012}, the L2 Syntactic Complexity Analyzer (L2SCA). There are nine syntactic measures : the number of words (W), the number of sentences (S), of verbal phrases (VP), of clauses (C), of T-units (T), dependent clauses (DC), complex T-units (CT), coordinated phrases (CP) and complex noun phrases (CN). The central unit for these metrics is the T-unit, defined as ‘one main clause plus any subordinate clause or nonclausal structure that is attached to or embedded in it’ (Hunt 1970:4).


Fourteen indices of syntactic complexity: the mean length of the sentence (MLS), the mean length of the T-unit (MLT), the mean length of the clause (MLC), the number of clauses per sentences (C/S), the number of verbal phrases per T-units (VP/T), the number of clauses per T-units (C/T), the number of dependent clauses per clauses (DC/C), the number of dependent clauses per T-units (DC/T).

\subsection{Readability metrics }
Readability metrics have traditionally been used to assess the difficulty of texts, i.e, how comprehensible or "readable" a text is for a particular audience. In that sense, for example, some of the metrics were initially designed to determine whether texts were suitable for particular school or college years, such as the Dale and Chall formula or the Bormuth's formulae. 

Most of the metrics take into account the average number of words per sentence as well as the average word length, such as the Automatic Readability Index (\textbf{ARI} = words per sentence + 9 $\cdot$ word length), or the \textbf{LIX} (number of words per sentence + percentage of words with more of 6 characters)  and \textbf{RIX} (number of long words / number of sentences) formulae; whereas other metrics also include syllable count, such as the well-known \textbf{Flesch-Kincaid} (0.39 $\cdot$ average sentence length + 11.8 $\cdot$ average number of syllables per word - 15.59) formula , the \textbf{Fog }(0.4 $\cdot$ (average sentence length + number of words with more than two syllabes), the \textbf{FORCAST} (20 - number of one-syllable words / 10), and the \textbf{Linsear Write} (number of one-syllable words + 3 $\cdot$ number of sentences) indexes. A couple of metrics also take into account the 'complexity' of the vocabulary deployed in the text by including parameters related to the use of 'difficult' or 'hard' words as defined by word lists previously created on the basis of native use, such as the \textbf{Dale and Chall} formula (0.1579 $\cdot$ percentage of difficult words + (0.496 $\cdot$ average sentence length) + 3.6365), the \textbf{Spache} grade (0.141 $\cdot$ average sentence length + (0.086 $\cdot$ percentage of unfamiliar words) + 0.839), or \textbf{Bormuth}'s four formulae. 

\section{Learner classification models} \label{sect:models}

\subsection{Prediction task set-up}

\begin{table*}[t]
\centering
\begin{tabular}{l|l|l|r|r|r|r|r}
  \hline
Features & Model & Partition & A1=>A2 & A2=>B1 & B1=>B2 & B2=>C1 & C1=>C2 \\ 
  \hline
Metrics & GBT & train & 0.897 & 0.888 & 0.903 & 0.854 & 0.948 \\ 
  Metrics & GBT & test & 0.895 & 0.897 & \textbf{0.778} & \textbf{0.821} & \textbf{0.587} \\ \hline 
  Term Freq. & Elastic Net & train & 0.972 & 0.977 & 0.895 & 0.998 & 0.949 \\ 
  Term Freq. & Elastic Net & test & 0.865 & 0.847 & 0.686 & 0.629 & 0.550 \\ \hline
  Word Seq. & LSTM & train & 0.985 & 0.985 & 0.942 & 0.777 & 0.634 \\ 
  Word Seq. & LSTM & test & 0.863 & 0.824 & 0.548 & 0.525 & 0.535 \\ \hline
  Metrics+ & GBT & train & 0.931 & 0.899 & 0.927 & 0.860 & 0.853 \\ 
  Metrics+ & GBT & test & \textbf{0.916} & \textbf{0.904} & 0.753 & 0.746 & 0.558 \\ 
   \hline
\end{tabular}
  \caption{AUC Metrics for binary classification models across four methods.}
  \label{tab:auc}
\end{table*}

The language learner ability levels in the available training data
are highly skewed towards beginners, from $17,605$ samples of learners
at the A1 level to only $76$ at the C2 level. One way to account for
this in building a predictive classifier is to only compare adjacent
classes. Rather than a single multinomial model, therefore, we use
5 pairwise models. While motivated by computational necessity, this
approach makes sense from a linguistic perspective as well. The
most interesting distinguishing marks between learners should come
from adjacent classes. Also, in most practical applications the
utility of a predictive model comes from distinguishing between
subtle differences in proficiency.

A second challenge in establishing a reliable classification algorithm
comes from differences in the tasks given as prompts in the EFCAMDAT 
dataset. Each of the essays is a response to a given question prompt
from a chosen topic such as \textit{Attending a robotics conference}
or \textit{Covering a news story}. Each of the $128$ topics is provided
as a prompt to only one specific learner level. If a modeling dataset 
was built by randomly assigning documents to training and testing sets,
it is likely that models could be predicting the topics rather than 
language proficiency (particularly if specific word counts are used).
For example, we may find that use of the word `robot' is a good
predictor of a learner achieving C2 proficiency because only those 
learners were asked to write about a robotics conference. In order to
avoid these spurious predictors, we split the data into training and
testing sets such that all of the essays from specific topics are 
grouped together. Similar strategies are commonly used in authorship
prediction tasks to separate the effects of topical and stylistic
predictors.

\subsection{Readability metrics} \label{ssec:rmetric}

The first set of models use only the computed readability metrics. Gradient boosted trees
are used for this task because they are generally very good general-purpose algorithms and,
unlike linear models, are particularly robust to highly correlated inputs. Trees are also
ideal for metric inputs defined on a scale, such as \textit{readability}, that is primarily
designed as an ordinal measurement.
Experimentation of the training set led to the choice of
a max tree depth of $4$, learning rate $0.01$ and a total of $100$ trees. The AUC metrics
for training and testing sets are shown
in Table~\ref{tab:auc}. We see that the classification task generally becomes harder for 
distinguishing between more advanced learners. Partially this is the result of differences
being more subtle and difficult to pick up from a small snippet of text. Also, the amount of
training data decreases when working with the higher proficiency learners. The models for
B1=>B2 and C1=>C2 seem to be overfit to the training data, but the other three models produce
models with AUC values that generalize favorably to the test set.

The top five features from each gradient boosted tree are shown in Table~\ref{tab:tabLevel}
based on the sum of model gain taken from all nodes that use a given variable as a split point.
The two strongest variables across the first four models are \textit{wordtokens} (number of
unique terms used) and  \textit{wordtypes} (number of word inflections used).

\begin{table*}[t]
  \centering
  \begin{tabular}{|l|l|l|l|l|}
    \hline
    \textbf{A1=>A2} & \textbf{A2=>B1} & \textbf{B1=>B2} &
    \textbf{B2=>C1} &  \textbf{C1=>C2}\\ \hline
    wordtokens &  wordtypes & wordtokens & wordtokens & ndwerz\\ 
    W & W & W & adjv & DC.C \\ 
    svv1 & ls2 & DC.C & vs1 & slextypes \\ 
    wordtypes & MLC & vs2 & swordtypes & MLC \\ 
    MLS & CN & lextokens & W & lv \\ 
 DC.T & CN.T & ttr & CN.T & modv \\ \hline
  \end{tabular}
  \caption{Most relevant metrics for pairwise level distinctions}
  \label{tab:tabLevel}
\end{table*}

\subsection{Term frequencies matrix}

In the second set of models, we use the actual word frequencies to predict the language
level of a learner. To do this, we tokenised the raw input text for each
learner and constructed a term frequency matrix $X$ of word counts per document. Words
that were used in less than 2\% of the corpus were filtered out. A logistic elastic net
regression was used to learn a predictive model from these word counts. The elastic net
model is a generalized linear regression model with an extra penalty term on the
negative log-likelihood. 
The elastic net works well with datasets with a large number of columns,
such as a term frequency matrix, because the $\ell_1$-penalty causes many of the values of
$\widehat{\beta}$ to be equal to zero (known as a parsimonious model). The degree of penalization was set using cross-validation.

In Table~\ref{tab:auc} we see that the word-based model improves the training-set AUC scores
significantly but leads to less predictive models on the test set. This is largely due to
overfitting caused by learning the topics discussed by the learner categories rather than
stylistic features that would generalize across tasks. However, the words along do perform
relatively well for distinguishing the A1=>A2 and A2=>B2 tasks. The dominant features in 
these models are the use of prepositions and verb forms that may correlate with an increase in proficiency.

\subsection{Incorporating word order}

One distinguishing feature of advanced learners is their ability to construct complex sentences
and correctly use advanced features such as zero-relative clauses. It would therefore seem that
a model that takes into account word-order would perform better for learning prediction compared
to methods based on term-frequency matrices. In order to incorporate word-order into a model, we
applied an LSTM model --- a particular type of recurrent neural network popular in text analysis
--- to the four classification tasks.
Given the relatively small data size, and following experimentation
on the training set, we selected an LSTM model that uses an embedding layer with $128$-dimensions,
$32$ recurrent units, and a high degree of dropout (80\%). The network was trained using the 
stochastic gradient descent, early stopping, and the learning rate schedule based on the ADAM 
algorithm. Results in Table~\ref{tab:auc} show that the LSTM model performs similarly to the 
elastic net for the first two tasks but not quite as well for classifying the advanced learners.
So, while in theory word-order should improve the model, the inconsistent topics and small amount
of data stops this from appearing empirically in the classification model. 

\subsection{Custom metrics}

Our final set of models constructs three new sets of metrics to add the default readability metrics
and re-applies the whole set of metrics using the same gradient boosted trees models shown in
Section~\ref{ssec:rmetric}. The new metrics incorporate features found when performing automatic
part-of-speech and dependency extraction on the corpus, which were extracted using the R package
\textbf{cleanNLP}. Specifically, we recorded the number of times
that (1) each universal part of speech code was used in a text, (2) each universal dependency was used in a text, and (3) words were used from each Zipf-scale categories (a map of 
each English word into a 7 category set based on its frequency in
usage).
We choose these metrics based on perceived features missing in the current readability metrics and
as an attempt to dig deeper in the the three most influential base metrics: \textit{worktoken} and
\textit{wordtypes}. The changes in AUC scores are shown in Table~\ref{tab:auc}. We see that the 
new features do improve the first models, producing the best AUC across all feature sets, but
due to the small sample sizes introduces a modest degree of overfitting. It should be noted that a cascading architecture only yields a 70.34 \% acccuracy.





\section{Discussion and Conclusions} \label{Sect:discuss-conclude}
\label{sec:discussion}


Taking into account the topic of the essay to be written, our experiment confirms the observations that task-based corpora entail strong overfitting \cite{alexopoulou2017task}. Another issue that cropped up with this type of data is that we have not quite overcome the initial skewed distribution of the data. For any expert system that we would like to build to automatically classify learner texts into learner levels, we have to face the fact that in most learner corpora, there will be more beginner and intermediate data than advanced (not to say "expert") data. 
Our experiment also suggests that more metrics, and not only readability formulae, may help improve classification rate, which somehow confirms the results presented in (\cite{vajjala_improving_2012}), where a classification accuracy of 93.3\% was achieved with 46 metrics. Following on this lead it is tempting to resort to even more complex or detailed systems producing metrics, such as the whole range of metrics produced by the Common Text Analysis Platform \cite{chen2016ctap}.

\subsection{Analyzing POS-tags}
We have shown that some POS-tag patterns could improve the classification rate. We have only used the Universal Part Of Speech (UPOS) tagset, which simplifies the analysis. More subtle tagsets (such as the Penn Treebank tagset) would possibly yield more fine-grained results at the expense of precision (these tagsets include more tags and are therefore linguistically more relevant but more error-prone.
We have not included punctuation in the analysis, but it is likely to play a role in the way more advanced learners conceive information packaging and structure sentence initial segments. 
As an aside, we ran frequency inventories of UPOS-grams for each level, to investigate whether we could see any specific patterns.
Initial levels exhibited the sequence 
\textit{noun / punctuation / pronoun / verb}.
The typical interpretation of this sequence is a sentence ending with a noun (end weight principle) and sentences beginning with a pronoun as a subject. In contrast, more advanced levels also favored more elaborated sequences with adverbs and postpositions. It should be noted that tagsets do not distinguish between commas and full stops (both labeled "punct") so that the analysis mostly holds for full stops, as learners do not always abide by expected punctuation guidelines (commas, and semi-colons are underused by learners). One strategy could consist in re-annotating the data with a specific tag for 'comma' and for 'full stop'.This would allow us to retrieve more specific information structure strategies across levels of proficiency. For instance, we would then be able to investigate which grammatical categories are used to express focus after the 'full stop'. As to the potential investigation of the use of 'comma', clustering the 'comma' with its corresponding dependency structure may help
classifying C1/C2 users by means of the required 'comma' for appositive clauses, for instance. Because learner punctuation might be erratic (especially for online-based data collection), 4-grams of POS-tags involving punctuation have proved to be more robust than 3-grams. 

\subsection{L1-based features}
The experiment has used data from French learners, but the metrics used are supposed to be language-independent. We could try to refine the classification by using features based on potential errors made by a given population of learners, such as their native language (L1). In this respect, further research could resort to specific problems for French learners, such as the expression of definiteness. For example, we would expect French speakers to overuse \textit{the} and \textit{a little} whereas \textit{much, few,} and \textit{fewer} would be underused.

\subsection{Lexically-based features}
It should be pointed out that some metrics rely on tokenized data, whereas others are computed on the basis of the raw texts. This is of paramount importance when it comes to learners because their (sometimes alternative and variable) spelling may artificially inflate the number of tokens. An expert system for learner levels should take this into account, especially for beginner levels. 
Discourse-based metrics could be used, such as the number of repetitions, more frequent in the A1 group.  
Last, we have mostly considered frequency as a potential cue for the use of the lexicon by learners, but more subtle techniques could be used, such as word embeddings. 

\bibliographystyle{abbrv}
\bibliography{cap2018}

\end{document}